\documentclass[journal]{IEEEtran}
\ifCLASSINFOpdf
\else
\fi
\usepackage[pdftex]{graphicx}

\usepackage{listings}
\usepackage{xcolor}
\usepackage{url}

\begin{document}
%
\title{Democratizing Tabular Data Access with an Open-Source Synthetic-Data SDK}
%
%
%



\author{
    Ivona Krchova$^{\ast}$,
    Mariana Vargas Vieyra$^{\ast}$,
    Mario Scriminaci,
    Andrey Sidorenko\thanks{Corresponding author: andrey.sidorenko@mostly.ai}
    \\
    \IEEEauthorblockA{MOSTLY AI}
    \thanks{$^\ast$Equal contribution.}
}

\maketitle

\begin{abstract}
Machine learning development critically depends on access to high-quality data. However, increasing restrictions due to privacy, proprietary interests, and ethical concerns have created significant barriers to data accessibility. Synthetic data offers a viable solution by enabling safe, broad data usage without compromising sensitive information. This paper presents the MOSTLY AI Synthetic Data Software
Development Kit (SDK), an open-source toolkit designed specifically for synthesizing high-quality tabular data. The SDK integrates robust features such as differential privacy guarantees, fairness-aware data generation, and automated quality assurance into a flexible and accessible Python interface. Leveraging the TabularARGN autoregressive framework, the SDK supports diverse data types and complex multi-table and sequential datasets, delivering competitive performance with notable improvements in speed and usability. Currently deployed both as a cloud service and locally installable software, the SDK has seen rapid adoption, highlighting its practicality in addressing real-world data bottlenecks and promoting widespread data democratization.
\end{abstract}


%
\IEEEpeerreviewmaketitle


\section{Introduction}
\label{sec:intro}

\IEEEPARstart{T}{he} development of Machine Learning applications requires broad access to training data. This necessity has become more critical in recent years with the advent of Deep Learning, which requires large-scale datasets to effectively train models. At the same time, the volume of data produced has grown rapidly, as companies and institutions now routinely collect vast amounts of data from user interactions and deployed applications. However, this exponential growth has not translated into equivalent accessibility. Privacy regulations, proprietary restrictions, and ethical considerations increasingly restrict the availability and usage of real-world datasets for development. This phenomenon, referred to as the data bottleneck \cite{Shani_2023}, affects multiple domains and data modalities with tabular datasets being no exception.


In this context, synthetic data arises as a promising solution to fill the data availability gap, offering means to securely access high-quality datasets, guaranteeing their utility and minimizing the privacy risks associated with data sharing  \cite{drechsler2011synthetic, jordon2022synthetic, united2023synthetic, hu2023sokprivacypreservingdatasynthesis}.
Synthetic datasets emerge as a versatile tool, also providing solutions for numerous problems other than privacy. 
For instance, scaling data volume, addressing class imbalance, imputing missing data or enabling fairness-aware evaluations \cite{NEURIPS2021_ba9fab00}.
Tackling these problems as a whole dramatically increases the usability and practicality of synthetic data \cite{vanbreugel2023privacynavigatingopportunitieschallenges}.
However, despite the growing body of literature leveraging Deep Learning for tabular data synthesis, many methods fall short of providing a holistic solution that addresses multiple aspects of data synthesis simultaneously, especially considering challenges like data heterogeneity, fidelity, and privacy protection \cite{bauer2024,shi2025comprehensivesurveysynthetictabular}.


To bridge the widening gap between data availability and accessibility, we introduce the MOSTLY AI Synthetic Data Software Development Kit (SDK), an open-source toolkit that integrates a comprehensive set of functionalities, simultaneously addressing multiple critical aspects of synthetic data generation.
This library is specifically tailored to reach a broad audience of researchers and practitioners of multiple fields, with the aim of filling the current data availability gap and promoting data democratization.
Designed for flexibility, the SDK can run locally or in `client mode', by accessing our custom-developed cloud-based service via API calls.
However, in this work we fully focus on the local deployment scenario, presenting its data ingestion, training, and generation capabilities in depth.


\subsection{Contributions}

The SDK offers the following key contributions:
\begin{itemize}
    \item \textit{High data quality with a simple design:} Based on TabularARGN \cite{tabularargn}, the SDK delivers synthetic data quality on a par with state-of-the-art (SOTA) models without compromising simplicity. Additionally, it achieves substantial speed-ups in training and generation times over current benchmarks.
    \item \textit{Privacy and fairness features:} The framework considers privacy-preserving value ranges for sampling, has built-in regularization layers and early stopping for training, and provides training mechanisms with differencial privacy (DP) guarantees \cite{dwo14-dp}. In generation, it supports the incorporation of fairness across sensitive attributes \cite{krc23}. 
    \item \textit{Automated quality assurance:} The SDK provides a built-in quality assurance module \cite{qareport}, generating standardized, in-depth reports that quantitatively assess both privacy and data fidelity of synthetic datasets.
    \item \textit{Open source package:} Fully written in Python, the SDK can be locally installed via \verb|pip|. The package is open-sourced under a fully permissive Apache Version 2.0 Open Source license\footnote{ \url{https://github.com/mostly-ai/mostlyai}}. 
    \item \textit{Community of users:} With a growing community of users and contributors, the SDK not only provides high-quality data synthesis services but also benefits from real-world feedback, enabling a continuous improvement of the toolkit to better meet real-world needs.
\end{itemize}

The remainder of the paper is organized as follows: Section II describes the SDK architecture; Section III presents data ingestion and processing capabilities; Section IV details training and generation capabilities; Section V reports empirical performance results; Section VI discusses community adoption; and Section VII concludes.


\section{Architecture Overview and Workflow}
\label{sec:sdk}

In this Section we present an overview of the SDK.
Built to support diverse usage scenarios, the SDK enables users from diverse backgrounds, from academic researchers to data science practitioners, to programmatically generate, share, and manipulate high-quality synthetic datasets. 
The toolkit provides a clearly structured interface.
It is freely available as an open-source package under the permissive Apache Version 2.0 license, 
and can be easily installed via \verb|pip install -U 'mostlyai[local]'| (requiring Python $\ge 3.10$).

\begin{figure*} 
     \centering
     \includegraphics[width=.9\textwidth]{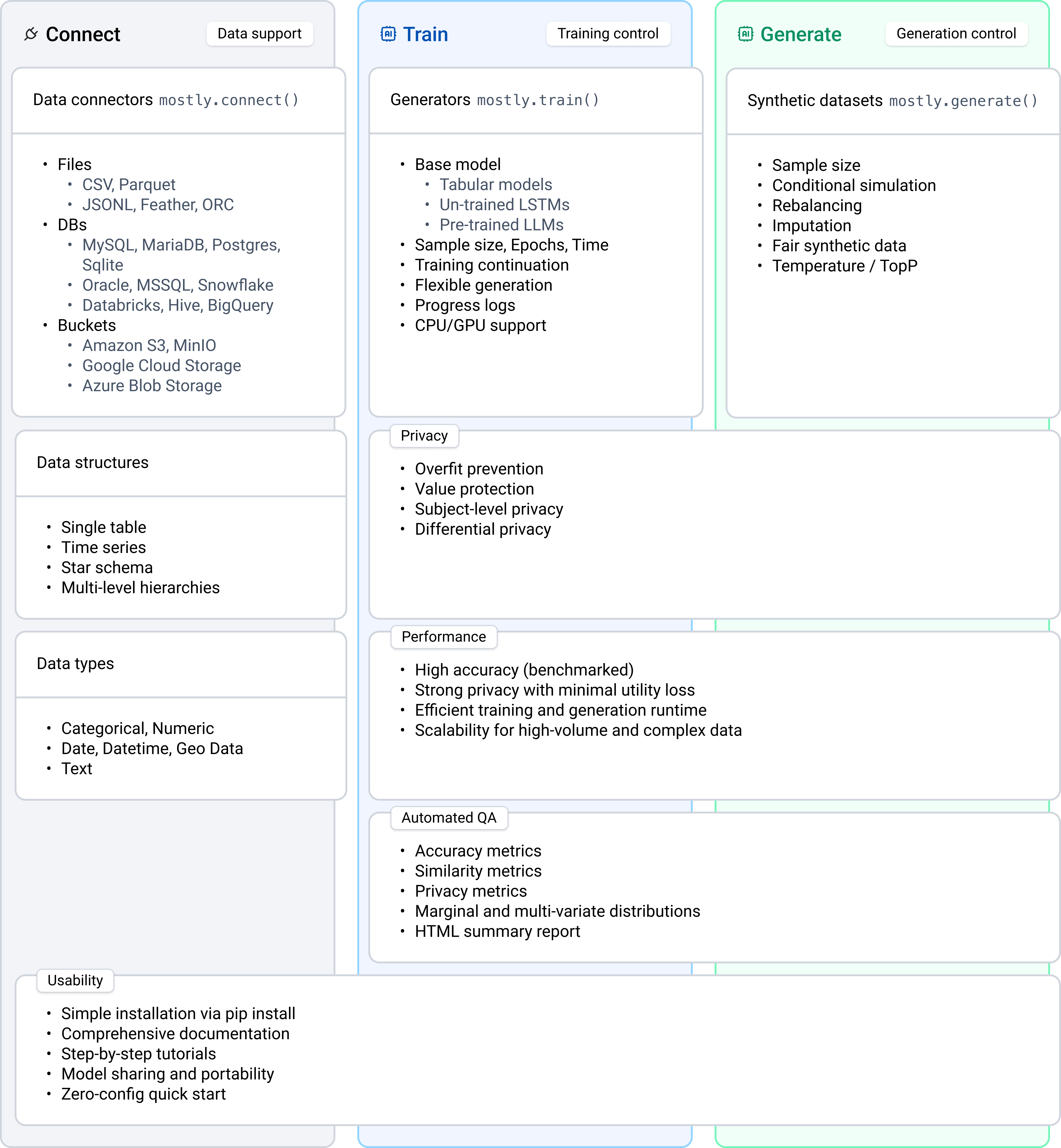}
     \caption{Overview of the key features and capabilities of the MOSTLY AI Synthetic Data SDK. The SDK is organized around three main artifacts, \emph{connectors}, \emph{generators} and \emph{synthetic datasets}. Connectors are responsible for the data support, including data ingestion and schema definition. Generators handle the training control and are mainly manipulated through the \texttt{mostly.train()} function. The generation is controlled by the synthetic dataset artifacts through the \texttt{mostly.generate()} function. Finally, the SDK integrates privacy mechanisms to ensure synthetic data safety, and a quality assurance toolkit to assess the quality of the resulting models and datasets.}
     \label{fig:features}
\end{figure*}

The SDK's architecture is organized around three main components:  \emph{connectors}, \emph{generators} and \emph{synthetic datasets}, alongside a dedicated \emph{quality assurance} component.
Figure~\ref{fig:features} illustrates this pipeline, highlighting the key functionalities associated with each component.
Together, these artifacts form an intuitive workflow that guides users from data ingestion, through training and sharing generative models, to generating and evaluating synthetic datasets.
While the SDK supports both local and client mode, in this work we focus on the local mode only, as both share the same functionalities and features.

In the following, we first introduce the main components and quality assurance in detail. 
Then a concise workflow example is presented to illustrate the practical usage of the SDK.

\subsection{Core Components of the SDK}
\label{subsec:components}

\paragraph{Connectors}
As the first step in the pipeline, the SDK provides interfaces for configuring data sources and destinations of various types.
These are managed via \emph{connectors}, which are containers that encapsulate all the information needed to access data sources for fetching training sets, or to specify destinations for storing the generated synthetic datasets.

Connectors are artifacts that enable a straightforward integration with various popular databases and cloud providers. 
In Figure~\ref{fig:features} we list the different types of connectors the SDK currently supports.
At the simplest level, a connector might only specify a local data path to a file the user uploaded manually, while more complex connectors can include credentials for authentication such as keyfiles, passwords, and SSL certificates.


\paragraph{Generators}
Once the training data has been fetched, the next step involves training a generative model.
This step is handled by a \emph{generator}, an artifact responsible not only for the training of the model itself but also for encapsulating all the relevant information about the dataset required to subsequently generate synthetic data.

Generators include functionalities such as support for multiple data types, integration of Large Language Models (LLMs) for text modeling, and configurable training settings. 
It also handles time-series data and multiple data topologies.
Moreover, the user can import and export generators for further development.

\paragraph{Synthetic Datasets}
The \emph{synthetic dataset} component handles the final step in the pipeline, which consists of generating synthetic data with the trained generator.
It contains the generated data, along with several additional artifacts, including data insights and quality assessment of synthetic data.
It also provides the full configuration details to facilitate reproducibility. 
Users can generate multiple datasets from a single trained generator, with options to resample records and generate data with different strategies.

\paragraph{Synthetic Data Quality Assurance Report}
The SDK provides an Automated Quality Assurance (QA) report for every synthetic data generator, enabling users to evaluate both fidelity and privacy of generated synthetic data. Each synthetic data generator is accompanied by a detailed QA report, which provides both standardized quality metrics and interactive HTML visualizations for comprehensive, interpretable analysis.

The evaluation framework \cite{qareport} uses a holdout-based approach to compare synthetic and original data, covering a range of quantitative metrics such as distributional similarity, embedding-based measures, and nearest-neighbor distances. It supports diverse data types, including sequential and contextual features. 

\subsection{Basic workflow}

The SDK aims to reach a wide audience by democratizing access to high-quality synthetic data, thus empowering researchers and practitioners. It is specifically designed for ease of use, clarity, and operational efficiency.

Figure~\ref{lst:snippet_1} shows a minimal end-to-end workflow for generating synthetic data. First, the SDK is initialized (line 3) and, as an example, the Titanic dataset is loaded into a \verb|Pandas| DataFrame (lines 5-7). A generator is then trained on this data using the \verb|mostly.train()| method (lines 8-11). The workflow concludes by displaying the QA report (line 12) before generating and accessing the final synthetic dataset (lines 13-14).

\begin{figure}%
\begin{lstlisting}[language=Python]
import pandas as pd
from mostlyai.sdk import MostlyAI
mostly = MostlyAI()

original_df = pd.read_csv(
  "https://github.com/mostly-ai/public-demo-data/raw/dev/titanic/titanic.csv"
)
g = mostly.train(
  name="QuickStartDemoTitanic",
  data=original_df,
)
g.reports(display=True)
sd = mostly.generate(g)
df = sd.data()
\end{lstlisting}
\caption{Basic workflow example}%
\label{lst:snippet_1}%
\end{figure}

After training on original data, a generator can be exported and shared, enabling others to generate synthetic data and adjust generation settings to their own requirements without exposing original data.
This feature is shown in Figure~\ref{lst:snippet_2}.
Once a generator is created, it is assigned a unique ID.
A user with permissions to access a specific generator can load the model weights as shown in line 1.
The SDK also allows to import and export generators as \verb|ZIP| files, as shown in lines 2 and 3.

\begin{figure}%
\begin{lstlisting}[language=Python]
g = mostly.generators.get('GENERATORID')
path_to_generator = g.export_to_file()
g = mostly.generators.import_from_file(path_to_generator)
\end{lstlisting}
\caption{Generator sharing}%
\label{lst:snippet_2}%
\end{figure}


\section{Data Ingestion and Processing}
\label{sec:data_ingestion}

Real-world datasets often come in diverse sizes and shapes, span multiple data modalities, and frequently comprise multiple tables with complex relationships. 
The SDK is precisely designed to handle this diversity, providing effective and robust mechanisms to accommodate different data modalities and schemas.
In this Section we first elaborate on how the SDK supports a wide range of data types through specialized strategies.
We then provide details about the supported data schemas.

\subsection{Data Preparation}
\label{subsec:data_encoding}



Before training, raw input data undergoes a comprehensive pre-processing pipeline that includes automatic type detection, encoding assignment, and privacy-preserving value protection mechanisms. 
The SDK follows the data encoding protocol proposed by \cite{tabularargn}, 
where all columns in the original dataset are transformed into one or more categorical sub-columns.
As a result, the encoded dataset exclusively consists of categorical variables, regardless of the domain of the original dataset. Table~\ref{table:supported_encodings} summarizes all the supported tabular encoding types.

For text features, a distinct modeling strategy is used. Text columns are handled separately, allowing users to first select the type of language model to apply. By default, our custom tokenizer is used, and an LSTM-based sequence model is trained from scratch. This model captures token sequences and their co-occurrences while incorporating contextual information from tabular columns. Alternatively, users may select a pre-trained LLM, in which case the model’s tokenizer is applied to tokenize the text data. The chosen pre-trained LLM is then fine-tuned on the dataset using Low-Rank Adaptation (LoRA), again integrating additional context from the tabular features into the text generation process.

Missing values are handled automatically as a separate category, which preserves their correlations with other features in the resulting synthetic data. Value protection mechanisms are applied to each column (see Table ~\ref{table:privacy_strategies}) ensuring that sensitive or outlier information is excluded from the synthetic data. 


\begin{table}[!t]
\renewcommand{\arraystretch}{1.3}
\centering
\caption{Tabular encoding types currently supported by the SDK.}
\label{table:supported_encodings}
\begin{tabular}{|c||p{0.63\linewidth}|}
    \hline
    \textbf{Encoding Type} & \textbf{Description} \\
    \hline
    Categorical & Encodes variables with a fixed set of predefined values (e.g., T-shirt sizes).\\
    \hline
    Numeric: Auto & Automatically selects the best numeric encoding strategy based on data heuristics. \\
    \hline
    Numeric: Discrete & Treats numeric data as categorical (e.g., ZIP codes, binary values). \\
    \hline
    Numeric: Binned & Bins numeric values into 100 discrete intervals, sampling within each bin to generate synthetic values.\\
    \hline
    Numeric: Digit & Splits numeric values into several categorical columns representing each digit in the number. \\
    \hline
    Character & Splits short strings with consistent formatting patterns into several categorical columns representing each character in the string. \\
    \hline
    Datetime & Splits datetime values into categorical sub-columns representing year, month, day, hour, minute and second. \\
    \hline
    Datetime: Relative & Models precise intervals between sequential events.\\
    \hline
    Latitude, Longitude & Converts latitude/longitude pairs into hierarchical quadtiles and encodes them as categorical sub-columns representing a progressively finer level of resolution.\\
    \hline
\end{tabular}
\end{table}

\begin{table}[t]
\renewcommand{\arraystretch}{1.3}
\centering
\caption{Summary of data-type-specific value protection strategies.}
\label{table:privacy_strategies}
\begin{tabular}{|c||p{0.7\linewidth}|}
\hline
\textbf{Column Type} & \textbf{Privacy Protection Strategy} \\
\hline
Categorical & Reduce re-identification risk from unique values by either consolidating these into a "\_RARE\_" category or by sampling replacement values from more common categories. \\
\hline
Numerical & Bin values into percentiles or decompose into digits to obscure exact numbers and clip the values to limit outlier identification. \\
\hline
Datetime & Decompose timestamps (year, month, day) and clip ranges to mask uncommon dates. \\
\hline
Geospatial & Convert latitude-longitude pairs into quadtiles with resolution adapted to local data density. \\
\hline
Text & Apply standard tokenizer with data-independent vocabulary to avoid phrase leakage.\\
\hline
\end{tabular}
\end{table}





\subsection{Managing Multi-table Relationships}

The SDK organizes datasets comprising multiple tables with two main types of table, \emph{subject tables} and \emph{linked tables}.
Subject tables contain exactly one record per subject, and define the level of granularity at which privacy mechanisms will be applied at training time. 
Linked tables, on the other hand, consist of records that represent events, sequences, or time-indexed data associated with subjects. 

Once a dataset is processed into the SDK, it can potentially consist of several tables of both types, with relationships between them established through two types of foreign keys: \emph{context foreign keys} and \emph{non-context foreign keys}.
When a subject and a linked table are connected through context foreign keys, the implemented algorithm ensures that the correlations between records from both tables are modeled and retained during generation.
In contrast, when a non-context foreign key is used, the referential integrity of the two connected tables is maintained, although they are modeled independently.
The toolkit uses these building blocks to connect tables within a given dataset, handling arbitrary data schemas with complex hierarchies and nested relationships.

\section{Generative Model Training}

Once the dataset is loaded and data schemas are defined, the SDK proceeds with model training.
At this stage the SDK automatically selects batch size and learning rate based on dataset size and available memory, and further optimizes convergence using a learning rate scheduler.
Users can leverage both CPU and GPU hardware to accelerate training, and real-time progress monitoring is provided for transparent tracking and management of training runs. 

In what follows, first an overview of the core model approach is provided, then details on the supported LLMs are given. Finally privacy preserving mechanisms implemented in the SDK are presented.

\subsection{Auto-regressive Tabular Model Overview}

The generator artifact is trained using the Tabular Auto-Regressive Generative Network (TabularARGN), detailed in~\cite{tabularargn}. 
Its architecture provides a shallow, any-order auto-regressive model that operates on discretized features and is optimized using categorical cross-entropy loss.
At its core, TabularARGN maps the generation process to an auto-regressive framework, estimating the joint probability distribution of the data as a product of simpler conditional probabilities. 
To handle multi-table datasets, the model extends the auto-regressive conditional probability framework to capture dependencies across related tables defined by the data schema.
In these scenarios, the SDK explicitly defines an ordering in which tables are processed, establishing parent-children relationships between them. 
This way, records from parent tables are compressed into \emph{context} embeddings, which are then used for conditional generation of their children tables.

A crucial aspect of TabularARGN is the order in which columns within a table are modeled.
In the SDK, when calling \texttt{mostly.train()}, users can choose between two column ordering strategies for training: fixed-order and flexible-order (or any-order). 
By default, the flexible option is enabled. In this mode, the SDK randomly shuffles the order of the columns for each training batch. This approach enables the model to learn how to generate data conditioned on any subset of features, making it versatile for tasks such as data imputation, conditional simulation, or fairness adjustments.
While flexible training allows for broader generation capabilities, it also makes the learning task more complex. This additional complexity can sometimes reduce the accuracy of the resulting synthetic data. Choosing a fixed column order simplifies training and can improve accuracy in some cases. However, once trained in fixed-order mode, the model loses the ability to generate data flexibly from arbitrary subsets of features.
This presents a trade-off between higher accuracy or more flexibility at generation time, with the optimal choice depending on the intended application.

\subsection{Support for LLM Finetuning}

We provide support for finetuning LLMs hosted in the HuggingFace platform.
As discussed in Section~2, one may opt to leverage a pre-trained LLM for modeling text columns. In this case, the user selects the desired LLM to be fine-tuned on the data, rather than relying on the default LSTM model, which is trained from scratch. This option is available in the SDK  as long as the user has a HuggingFace access token with read permissions.
We currently support any text-generation LLMs available on HuggingFace.

\subsection{Privacy Mechanisms}

The TabularARGN training procedure is designed with strong privacy safeguards to protect individual records in the training data. To minimize the risk of overfitting, which can lead to the memorization of sensitive information, the model employs dedicated privacy-oriented strategies:
\begin{itemize}
    \item \textbf{Early Stopping:} During training, up to 10\% of the dataset is set aside as a validation set. At the end of each epoch, validation loss is calculated. If the validation loss does not improve for $N$ consecutive epochs (default: 5), training stops early. The model weights corresponding to the lowest observed validation loss are retained as the final model.
    \item \textbf{Dropout Regularization:}  Each network layer includes a dropout rate of 25\%, reducing the risk of overfitting by randomly deactivating a subset of neurons during training.
\end{itemize}

\paragraph{Differential Privacy}
The SDK offers the option to train synthetic data generators with DP guarantees. When DP is enabled, the training procedure ensures that individual records in the original dataset have a mathematically bounded influence on the trained model, as quantified by the privacy budget parameter $\epsilon$. The privacy budget is monitored throughout training, and an upper $\epsilon$ limit can be set to automatically halt training once this threshold is reached.

Differential privacy in the SDK is implemented using the Opacus library, which enforces privacy guarantees through gradient clipping, noise addition, randomized data loading, and privacy accounting. Users can configure the level of noise and the clipping parameters to balance privacy protection and model utility according to their requirements.

While non-DP training is the default to maximize efficiency and accuracy, the DP option is available for users who require strict mathematical privacy guarantees, with the understanding that this may impact model performance and computational efficiency.


\section{Synthetic Data Generation}
Once training is complete, the generator produces synthetic data that matches the statistical properties of the original dataset. 
Synthetic data generation is performed with the \texttt{mostly.generate()} function, which uses the trained generator as input.


In setups involving two related tables (e.g., a flat subject table and a sequential linked table), generation proceeds hierarchically. The flat table is sampled first, providing the context required by the sequential table model. The sequential table is then generated conditionally, leveraging the synthetic records from the flat table as contextual input.

\paragraph{Multi-Table Generation}
For multi-table schemas, the generation process is governed by the dependencies and contextual relationships learned during training. The schema hierarchy dictates the sampling order: parent and ancestor tables are synthesized first, providing context for all downstream tables. When generating a target table, the process conditions on synthetic values from all tables in the context chain, as well as one-hop context tables. For sibling sequential tables, the SDK supports conditional generation, allowing the output of one sequence to be used as context for another and thereby capturing correlations between related sequences. Tables not included in the context chain or as one-hop context are generated independently. However, referential integrity is strictly maintained: all foreign key relationships, even for tables more than one hop away, are preserved in the synthetic dataset. This ensures that all links between records are valid, upholding the structural consistency required for realistic multi-table data.

\subsection{Flexible Generation}
Flexible generation allows the model to go beyond simply reproducing the statistical patterns of the training data. This feature, available for models that have been trained with a flexible-order column ordering, includes the ability to generate any desired amount of synthetic data, from small samples to large-scale datasets. Flexible conditional simulation enables users to provide specific values for selected columns, with synthetic content generated for the remaining columns, making it possible to tailor outputs to specific scenarios. For a variety of advanced transformations (such as rebalancing, imputation, and fairness adjustment) users only need to specify the target columns and the SDK automatically manages the optimal column ordering for each scenario. 


\paragraph{Conditional Simulation}
Conditional simulation empowers users to move beyond simple predictions by generating multiple synthetic scenarios conditioned on initial attributes. Instead of providing just a single expected value, the model produces an empirical distribution that quantifies the range and likelihood of possible outcomes. For instance, in healthcare analytics, one can simulate patient recovery trajectories under different treatment plans, using available patient data as context and then sampling plausible recovery scenarios. This flexible approach offers deeper insight into uncertainty and possible futures, supporting more informed, data-driven decisions.

\paragraph{Rebalancing}
Rebalancing enables users to address class imbalance by generating additional synthetic samples for minority groups in the data. By specifying desired class proportions, users can increase the representation of underrepresented outcomes while ensuring that the remaining features remain consistent and realistic. In comparative evaluations \cite{upsampling}, TabularARGN-based upsampling  has demonstrated advantages over traditional techniques such as naive oversampling \cite{imblearn_naive} and SMOTE-NC \cite{imblearn_smotenc}. Unlike these methods, which often duplicate existing records or interpolate between them, TabularARGN generates a diverse set of realistic synthetic samples that more accurately reflect the underlying data distribution, leading to better performance for machine learning model trained on these balanced datasets, particularly when the minority class is very small \footnote{See the SDK usage tutorial for instructions on rebalancing: \url{https://github.com/mostly-ai/mostlyai/tree/main/docs/tutorials/rebalancing}}.

\paragraph{Imputation}
Imputation enables users to fill in missing values in their data by leveraging the statistical relationships learned during training. For example, if a dataset is missing values in a categorical column such as occupation, the generator can use available features such as education, age, and income to infer plausible occupations for those individuals.  Instead of replicating missing patterns, the generator masks missing categories and samples plausible values, positioning imputed columns last to use all available context. For conditional imputation, users provide a seed with missing entries and observed features as context; SDK then fills in missing values accordingly
\footnote{For detailed guidance on using the rebalancing feature, please refer to the SDK usage tutorial at \url{https://github.com/mostly-ai/mostlyai/tree/main/docs/tutorials/smart-imputation}.}.

\paragraph{Fairness}
Fairness adjustment enables users to generate synthetic data that satisfies the statistical parity criterion, ensuring balanced outcomes across sensitive groups. For instance, in the original data, females may be underrepresented among high-income people, resulting in unfair downstream model predictions. The fairness adjustment feature in the SDK modifies the data generation process so that sensitive groups have equal chances of being assigned a positive outcome, regardless of existing bias in the original dataset.

This is achieved by sampling the target column last and adjusting its conditional probabilities based on the previously generated sensitive attributes, thus enforcing statistical parity. As a result, users can create fairness-aware datasets, allowing downstream models to produce fairer, more balanced predictions, even when evaluated on real, biased data. These constraints are applied dynamically during generation, making it possible to flexibly manage the trade-off between fairness and accuracy without retraining the generator~\cite{krc23}\footnote{To learn more about using the fairness feature, check out the SDK tutorial at \url{https://github.com/mostly-ai/mostlyai/tree/main/docs/tutorials/fairness}.}.


\paragraph{Flexible Sampling}
In addition to flexible generation order, the engine allows users to further control the sampling process by adjusting the temperature and top-p parameters. Lowering the temperature (e.g. $< 1$) makes the model more likely to generate common values, which is useful when users want output that strictly follows known rules or business logic, for example, generating only valid product codes or adhering closely to typical user behaviors. The rise in temperature (e.g. $> 1$) increases the variety of generated data, which can be valuable for stress testing models or exploring rare scenarios.

The top-p parameter sets a threshold for the cumulative probability mass considered during sampling; setting a lower top-p value focuses generation on the most probable outcomes, while a higher value allows more diversity by including less likely options. Together, these controls let users fine-tune the diversity and representativeness of synthetic data to match specific needs, such as creating realistic records for routine operations or generating edge cases for robust model evaluation.


\section{Empirical Performance and Community Adoption}
\label{sec:usage}
\begin{figure*}
\centering
\includegraphics[width=.9\textwidth]{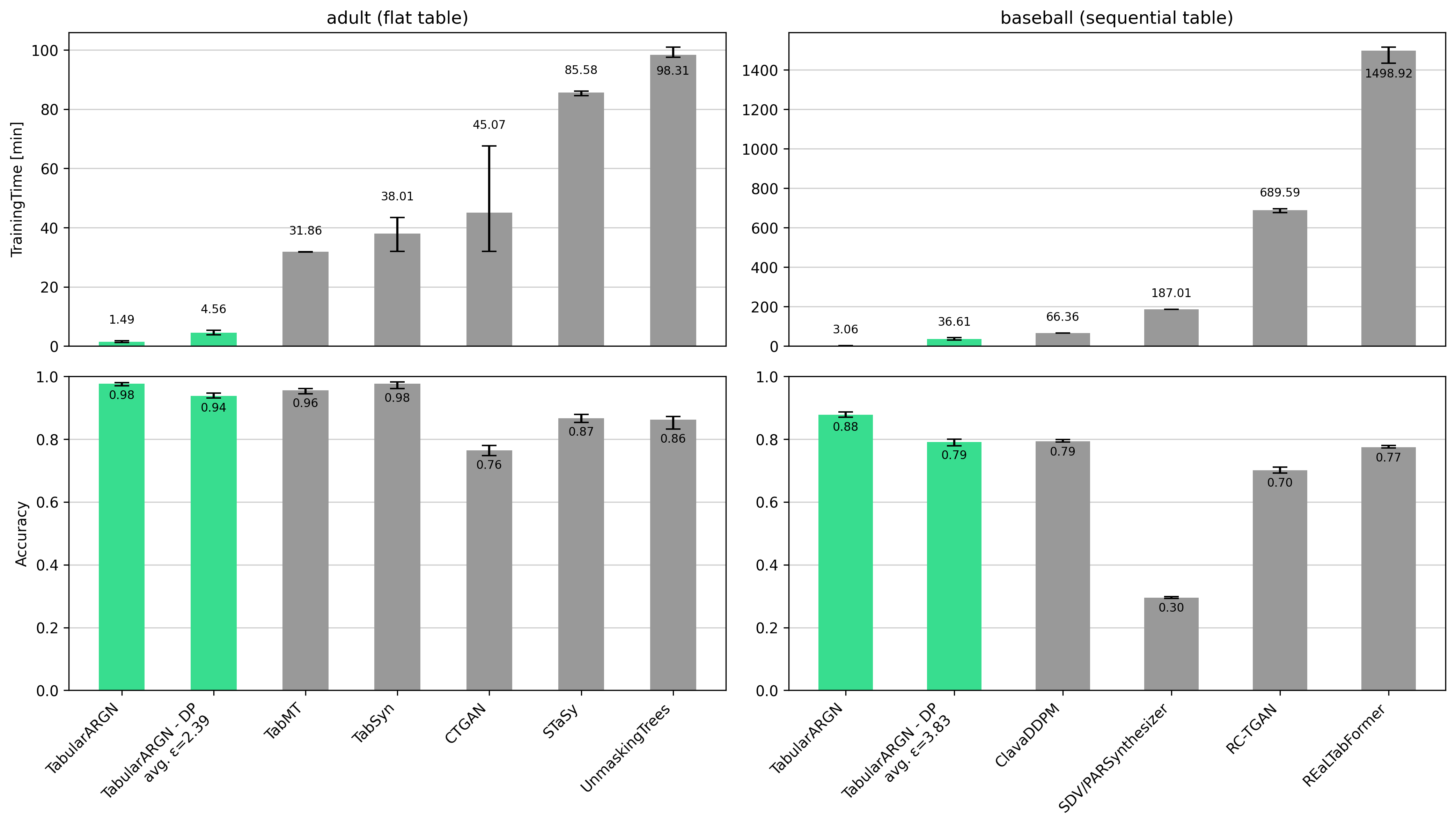}
\caption{Training time (top) and accuracy (bottom) for the flat \emph{Adult} (left) and sequential \emph{Baseball} (right) datasets. Values are averaged over three or more runs; error bars indicate min/max. TabularARGN is shown both with and without DP training.}
\label{fig:results}
\end{figure*}

\begin{figure}[t]
    \centering
    \includegraphics[width=\columnwidth]{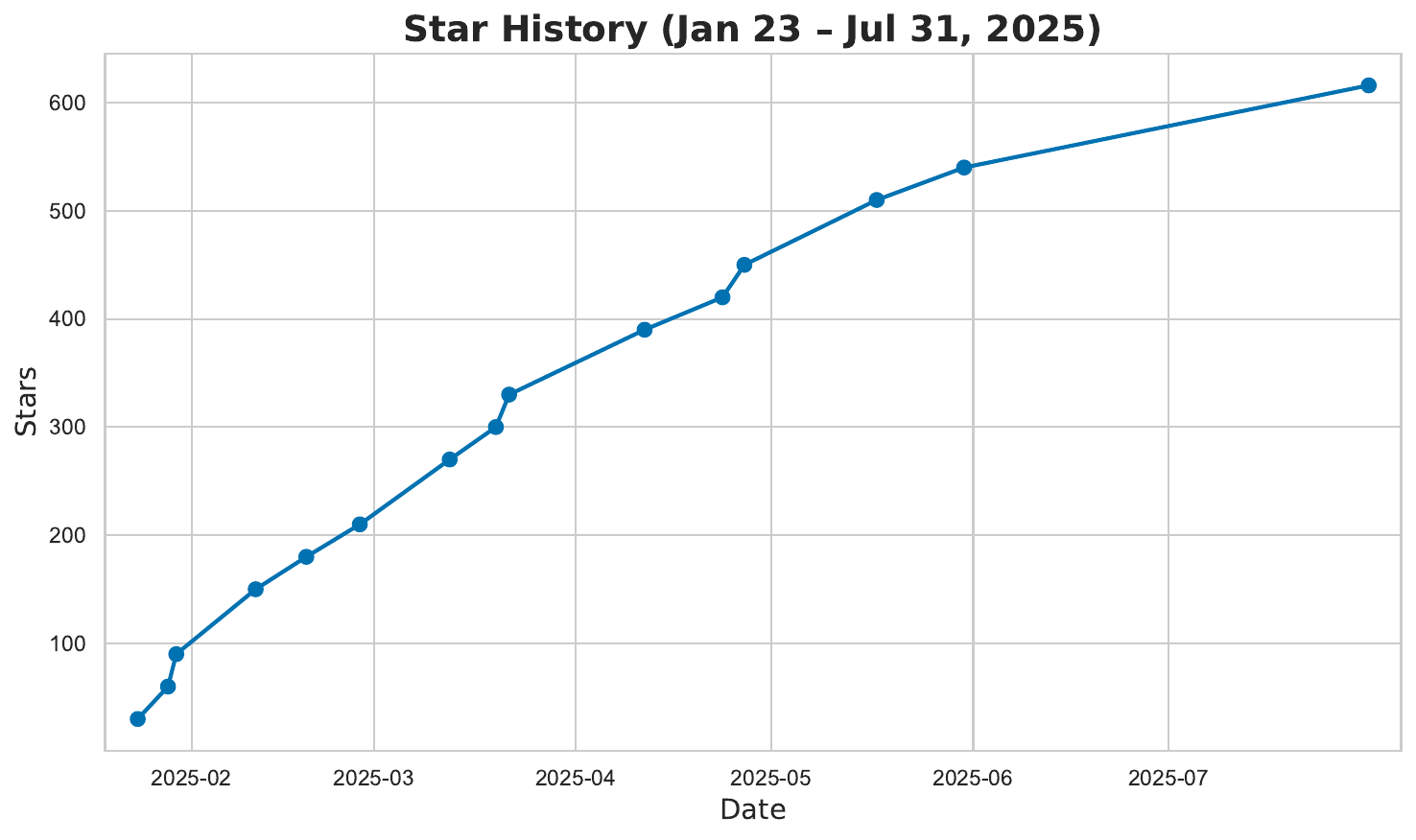}
    \caption{GitHub stars history for the SDK since its release on January 23rd 2025 until July 31st 2025.}
    \label{fig:sdk-star-history}
\end{figure}

\begin{figure}[t]
\centering
\includegraphics[width=.9\columnwidth]{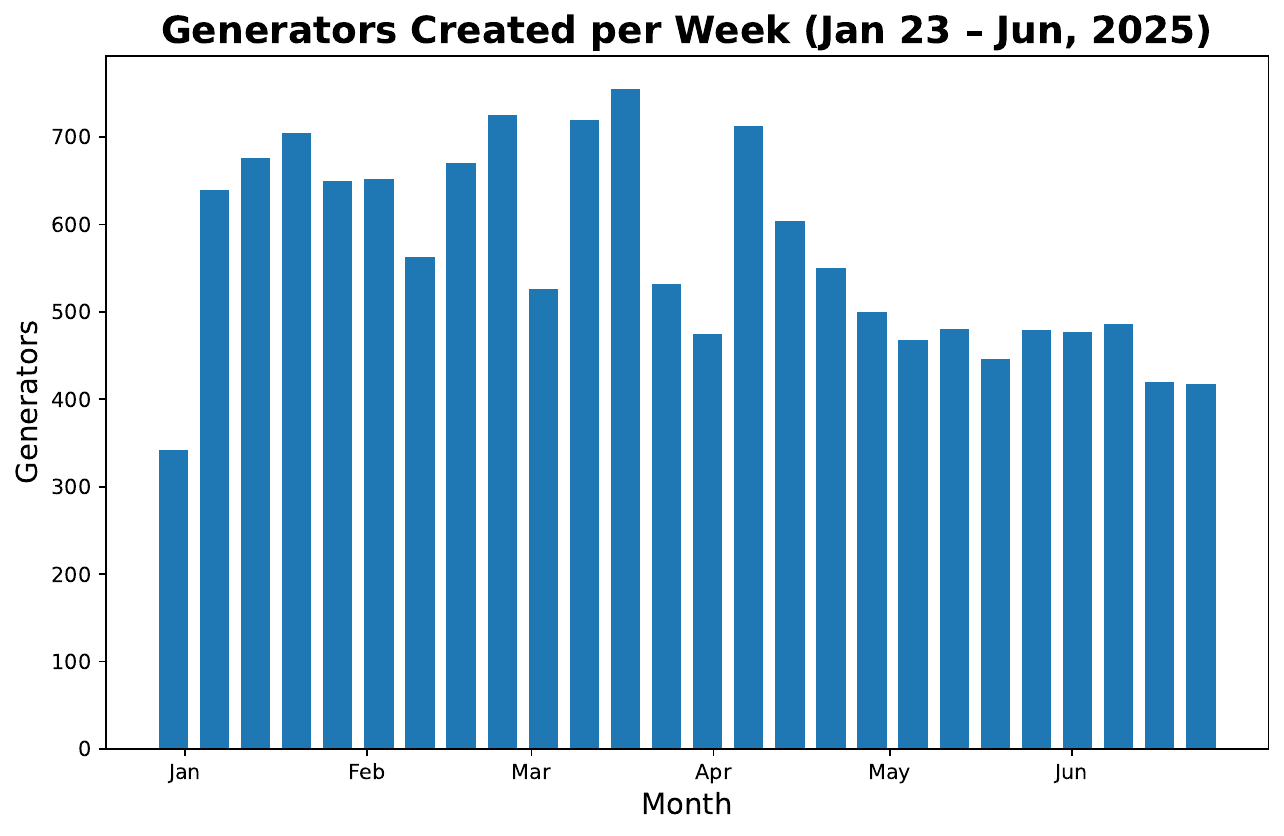} 
\caption{Number of generators per week created by users external to the MOSTLY AI organization since the release date up to June.}
\label{fig:nb_generators}
\end{figure}

The SDK offers a practical solution for a variety of data challenges, including privacy-safe synthesis, rebalancing of skewed datasets, imputation of missing values, and fair synthetic data.
To comprehensively assess the real-world impact of the SDK we elaborate on two complementary aspects of our work. 
On the one hand, the empirical evidence presented by \cite{tabularargn} is summarized, highlighting the strong performance of TabularARGN, the core model of the SDK, in terms of generation accuracy, robust privacy guarantees, and efficient runtime.
On the other hand, the growing community adoption observed since the SDK's public release in January 2025 is discussed, demonstating its practicality, usability, and relevance.

\subsection{Synthetic Data Quality}


To demonstrate the effectiveness of our approach, we summarize key empirical results previously reported in \cite{tabularargn}, where TabularARGN was benchmarked against state-of-the-art methods for both flat and sequential tabular datasets. 
Evaluations were carried out following the protocol proposed by \cite{pla21-qa}, which assesses the quality of synthetic data in terms of both statistical fidelity, measured through low-order marginal statistics such as univariate and bivariate accuracies, and privacy protection quality, quantified via metrics like Distance to Closest Record (DCR) share.
This evaluation protocol is implemented in an open-source package~\cite{qareport} available at \cite{mostlyai-qa}. 
The SDK leverages this package to produce the comprehensive quality assurance reports described previously in Section 2.

For flat tables, the model was compared with methods such as CTGAN \cite{xu19}, STaSy \cite{kim23-StaSy}, TabSyn \cite{zha24-TabSyn}, TabMT \cite{gul23-TabMT}, and Unmasking Trees \cite{mcc24-UnmaskingTrees}. For sequential data, benchmarks included REalTabFormer \cite{sol23-Realtabformer}, RC-TGAN \cite{gue23-rctgan}, and ClavaDDPM \cite{pan24-clavaddpm}.
Although several datasets were examined, in this Section we only present results of one dataset for each data regime, the \emph{adult} census dataset and the \emph{baseball} dataset, comprising longitudinal player-performance statistics.
Figure~\ref{fig:results} shows the main results, which underscore TabularARGN's ability to achieve state-of-the-art performance while providing significant speed-ups over the benchmarks, even when training with DP, which can negatively impact accuracy and runtime.

\subsection{Scalability and Complex Data Handling}
Despite significant progress in the field, the majority of deep learning–based tabular data generators remain focused on solving only subsets of problems, such as privacy or fidelity, often at the expense of flexibility, scalability, or support for real-world database topologies. A prominent example is the Synthetic Data Vault (SDV)~\cite{pat16-sdv}, a widely used Business Source Python library developed by DataCebo, which offers tools for single-table, multi-table, and time-series data synthesis, leveraging a range of generative modeling techniques including GANs and VAEs. However, the benchmarking and experimental assessments \footnote{SDV vs. SDK: Single table scenario at \url{https://mostly.ai/blog/a-comparison-of-synthetic-data-vault-and-mostly-ai-part-1-single-table-scenario}} show that while SDV excels in certain single-table use cases, it can encounter scalability and performance limitations with larger or more complex datasets, and may require manual configuration or extensive parameter tuning.
Extending this analysis to a two-table dataset with complex relational and temporal dependencies, we observed that SDV struggled to capture these patterns, whereas the SDK handled them more effectively, achieving higher fidelity and utility under these conditions \footnote{SDV vs. SDK: Sequential Scenario at \url{https://mostly.ai/blog/a-comparison-of-synthetic-data-vault-and-mostly-ai-part-2-sequential-scenario}} .

\subsection{Community Adoption}

Since its release on January 23rd 2025, the SDK has seen rapid adoption, with growing engagement on GitHub. Figure~\ref{fig:sdk-star-history} tracks the increase in GitHub stars from the release date up to July 31, a strong indicator of interest. 
Furthermore, in Figure~\ref{fig:nb_generators} we present the number of generators per week that have been created by users that do not belong to the MOSTLY AI organization.
Importantly, these figures only represent generators created in cloud service, and therefore serve as a lower bound of actual usage, which includes generators created locally. 
This underscores the substantial community engagement with the toolkit.

There is growing enthusiasm that is already driving practical applications. 
A recent example is a visualization tool for energy performance certificates developed by the Moderate Project, a European organization that aims to enable companies and institutions to extract actionable knowledge from data by providing data services and analytics\footnote{Moderate's main website: \url{https://moderate-project.eu/in-a-nutshell/}}.
The Moderate team integrated the SDK into their geo-clustering tool, making it possible to develop a visualization tool for analyzing energy performance certificates for the Piedmont region of Italy.
As the underlying data is not publicly available, they leverage synthetic data to make this tool available for public use.\footnote{Their visualization tool is available at \url{https://tools.eeb.eurac.edu/epc_clustering/piemonte/}.}

\paragraph{Further Resources}
Usage guidance and extensive tutorials covering all the main features of the SDKof the SDK are also provided to help users at all levels quickly get started with the SDK and explore its full capabilities \footnote{\url{https://mostly-ai.github.io/mostlyai/tutorials/}}.


\section{Conclusion}
\label{sec:conclusion}
This paper has introduced the architecture and key functionalities of the MOSTLY AI Synthetic Data SDK, focusing on how 
its intuitive interface equips the practitioner with a holistic solution to address a wide range of real-world, practical use cases.  
The SDK offers a scalable solution for privacy-preserving synthetic data generation, supporting a wide variety of tabular data topologies and mixed data types.

With its modular design and rich set of configuration options, the SDK enables users to address common data challenges such as class imbalance, missing values, and fairness requirements, all within an accessible and easy-to-use Python API. The ability to adjust key aspects of training and generation, combined with straightforward local deployment, makes privacy-compliant data access more accessible and accelerates the adoption of synthetic data in practice. By offering an open, locally deployable solution, the SDK also enables secure data-driven innovation. Future work will further enhance model capabilities and extend support for additional data modalities.

\section*{Acknowledgment}

We gratefully acknowledge Kerem Erdem, Michael Druk, Alex Ichim, André Jonasson, Lukasz Kolodziejczyk, Duygu Okcu, Dmitry Parshenkov, Michael Platzer, Radu Rogojanu, Paul Tiwald, and Shuang Wu for their invaluable contributions to the development of the SDK. Their collective expertise and dedication have played a pivotal role in advancing the SDK’s capabilities, ensuring high-quality synthetic data generation while maintaining rigorous safeguards against information leakage. Their efforts have been instrumental in making the SDK an efficient, robust, flexible, and user-friendly solution for tabular synthetic data synthesis.

\ifCLASSOPTIONcaptionsoff
  \newpage
\fi



\bibliographystyle{IEEEtran}
\bibliography{main}

@misc{vanbreugel2023privacynavigatingopportunitieschallenges,
      title={Beyond Privacy: Navigating the Opportunities and Challenges of Synthetic Data}, 
      author={Boris van Breugel and Mihaela van der Schaar},
      year={2023},
      eprint={2304.03722},
      archivePrefix={arXiv},
      primaryClass={cs.LG},
      url={https://arxiv.org/abs/2304.03722}, 
}

@inproceedings{NEURIPS2021_ba9fab00,
 author = {van der Schaar, Mihaela and van Breugel, Boris and Kyono, Trent and Berrevoets, Jeroen},
 booktitle = {Advances in Neural Information Processing Systems},
 editor = {M. Ranzato and A. Beygelzimer and Y. Dauphin and P.S. Liang and J. Wortman Vaughan},
 pages = {22221--22233},
 publisher = {Curran Associates, Inc.},
 title = {DECAF:  Generating Fair Synthetic Data Using Causally-Aware Generative Networks},
 volume = {34},
 year = {2021}
}

@article{xu19,
  title={Modeling tabular data using conditional gan},
  author={Xu, Lei and Skoularidou, Maria and Cuesta-Infante, Alfredo and Veeramachaneni, Kalyan},
  journal={Advances in neural information processing systems},
  volume={32},
  year={2019}
}

@inproceedings{gue23-rctgan,
  title={Row conditional-TGAN for generating synthetic relational databases},
  author={Gueye, Mohamed and Attabi, Yazid and Dumas, Maxime},
  booktitle={ICASSP 2023-2023 IEEE International Conference on Acoustics, Speech and Signal Processing (ICASSP)},
  pages={1--5},
  year={2023},
  organization={IEEE}
}

@misc{pan24-clavaddpm,
      title={ClavaDDPM: Multi-relational Data Synthesis with Cluster-guided Diffusion Models}, 
      author={Wei Pang and Masoumeh Shafieinejad and Lucy Liu and Stephanie Hazlewood and Xi He},
      year={2024},
      eprint={2405.17724},
      archivePrefix={arXiv},
      primaryClass={cs.AI},
      url={https://arxiv.org/abs/2405.17724}, 
}

@inproceedings{pat16-sdv,
    title={The Synthetic data vault},
    author={Patki, Neha and Wedge, Roy and Veeramachaneni, Kalyan},
    booktitle={IEEE International Conference on Data Science and Advanced Analytics (DSAA)},
    year={2016},
    pages={399-410},
    doi={10.1109/DSAA.2016.49},
    month={Oct}
}

@inproceedings{gul23-TabMT,
 author = {Gulati, Manbir and Roysdon, Paul},
 booktitle = {Advances in Neural Information Processing Systems},
 editor = {A. Oh and T. Naumann and A. Globerson and K. Saenko and M. Hardt and S. Levine},
 pages = {46245--46254},
 publisher = {Curran Associates, Inc.},
 title = {TabMT: Generating tabular data with masked transformers},
 volume = {36},
 year = {2023}
}

@article{pla21-qa,
  title={Holdout-based empirical assessment of mixed-type synthetic data},
  author={Platzer, Michael and Reutterer, Thomas},
  journal={Frontiers in big Data},
  volume={4},
  pages={679939},
  year={2021},
  publisher={Frontiers Media SA}
}

@article{sol23-Realtabformer,
  title={Realtabformer: Generating realistic relational and tabular data using transformers},
  author={Solatorio, Aivin V and Dupriez, Olivier},
  journal={arXiv preprint arXiv:2302.02041},
  year={2023}
}

@inproceedings{zha24-TabSyn,
  title={Mixed-Type Tabular Data Synthesis with Score-based Diffusion in Latent Space},
  author={Zhang, Hengrui and Zhang, Jiani and Srinivasan, Balasubramaniam and Shen, Zhengyuan and Qin, Xiao and Faloutsos, Christos and Rangwala, Huzefa and Karypis, George},
  booktitle={The Twelfth International Conference on Learning Representations},
  year={2024}
}

@inproceedings{kim23-StaSy,
  title={StaSy: Score-based tabular data synthesis}, 
  author={Kim, Jayoung and Lee, Chaejeong and Park, Noseong},
  booktitle={The Elenventh International Conference on Learning Representations},
  year={2023}
}

@article{mcc24-UnmaskingTrees,
  title={Unmasking trees for tabular data},
  author={McCarter, Calvin},
  journal={arXiv preprint arXiv:2407.05593},
  year={2024}
}

@misc{krc23,
      title={Strong statistical parity through fair synthetic data}, 
      author={Ivona Krchova and Michael Platzer and Paul Tiwald},
      year={2023},
      eprint={2311.03000},
      archivePrefix={arXiv},
      primaryClass={cs.LG},
      url={https://arxiv.org/abs/2311.03000}, 
}

@article{dwo14-dp,
  title={The algorithmic foundations of differential privacy},
  author={Dwork, Cynthia and Roth, Aaron and others},
  journal={Foundations and Trends{\textregistered} in Theoretical Computer Science},
  volume={9},
  number={3--4},
  pages={211--407},
  year={2014},
  publisher={Now Publishers, Inc.}
}

@article{jordon2022synthetic,
  title={Synthetic Data--what, why and how?},
  author={Jordon, James and Szpruch, Lukasz and Houssiau, Florimond and Bottarelli, Mirko and Cherubin, Giovanni and Maple, Carsten and Cohen, Samuel N and Weller, Adrian},
  journal={arXiv preprint arXiv:2205.03257},
  year={2022}
}

@book{united2023synthetic,
   author = {{UNECE}},
   title = {Synthetic Data for Official Statistics},
   publisher = {United Nations},
   year={2023},
   url = {https://www.un-ilibrary.org/content/books/9789210021708},
}

@book{drechsler2011synthetic,
  title={Synthetic datasets for statistical disclosure control: theory and implementation},
  author={Drechsler, J{\"o}rg},
  volume={201},
  year={2011},
  publisher={Springer Science \& Business Media}
}

@misc{hu2023sokprivacypreservingdatasynthesis,
      title={SoK: Privacy-Preserving Data Synthesis}, 
      author={Yuzheng Hu and Fan Wu and Qinbin Li and Yunhui Long and Gonzalo Munilla Garrido and Chang Ge and Bolin Ding and David Forsyth and Bo Li and Dawn Song},
      year={2023},
      eprint={2307.02106},
      archivePrefix={arXiv},
      primaryClass={cs.CR},
      url={https://arxiv.org/abs/2307.02106}, 
}

@software{mostlyai-qa,
    author = {{MOSTLY AI}},
    title = {{MOSTLY AI - Quality Assurance}},
    url = {https://github.com/mostly-ai/mostlyai-qa},
    year = {2024}
}

@misc{imblearn_naive,
  title = "{RandomOverSampler} Documentation for RandomOverSampler",
  author = "{scikit-learn}",
  howpublished = "\url{https://imbalanced-learn.org/stable/references/generated/imblearn.over_sampling.RandomOverSampler.html}",
  year = {2011},
}

@misc{imblearn_smotenc,
  title = "{SMOTENC} Documentation for SMOTENC",
  author = "{scikit-learn}",
  howpublished = "\url{https://imbalanced-learn.org/stable/references/generated/imblearn.over_sampling.SMOTENC.html}",
  year = {2011},

}

@article{Shani_2023,
   title={The Lean Data Scientist: Recent Advances Toward Overcoming the Data Bottleneck},
   volume={66},
   ISSN={1557-7317},
   url={http://dx.doi.org/10.1145/3551635},
   DOI={10.1145/3551635},
   number={2},
   journal={Communications of the ACM},
   publisher={Association for Computing Machinery (ACM)},
   author={Shani, Chen and Zarecki, Jonathan and Shahaf, Dafna},
   year={2023},
   month=jan, pages={92–102} }

@misc{tabularargn,
      title={TabularARGN: A Flexible and Efficient Auto-Regressive Framework for Generating High-Fidelity Synthetic Data}, 
      author={Paul Tiwald and Ivona Krchova and Andrey Sidorenko and Mariana Vargas Vieyra and Mario Scriminaci and Michael Platzer},
      year={2025},
      eprint={2501.12012},
      archivePrefix={arXiv},
      primaryClass={cs.LG},
      url={https://arxiv.org/abs/2501.12012}, 
}

@misc{qareport,
      title={Benchmarking Synthetic Tabular Data: A Multi-Dimensional Evaluation Framework}, 
      author={Andrey Sidorenko and Michael Platzer and Mario Scriminaci and Paul Tiwald},
      year={2025},
      eprint={2504.01908},
      archivePrefix={arXiv},
      primaryClass={cs.LG},
      url={https://arxiv.org/abs/2504.01908}, 
}

@misc{upsampling,
      title={Improving Predictions on Highly Unbalanced Data Using Open Source Synthetic Data Upsampling}, 
      author={Ivona Krchova and Michael Platzer and Paul Tiwald},
      year={2025},
      eprint={2507.16419},
      archivePrefix={arXiv},
      primaryClass={cs.LG},
      url={https://arxiv.org/abs/2507.16419}, 
}

@misc{bauer2024,
      title={Comprehensive Exploration of Synthetic Data Generation: A Survey}, 
      author={André Bauer and Simon Trapp and Michael Stenger and Robert Leppich and Samuel Kounev and Mark Leznik and Kyle Chard and Ian Foster},
      year={2024},
      eprint={2401.02524},
      archivePrefix={arXiv},
      primaryClass={cs.LG},
      url={https://arxiv.org/abs/2401.02524}, 
}

@misc{shi2025comprehensivesurveysynthetictabular,
      title={A Comprehensive Survey of Synthetic Tabular Data Generation}, 
      author={Ruxue Shi and Yili Wang and Mengnan Du and Xu Shen and Yi Chang and Xin Wang},
      year={2025},
      eprint={2504.16506},
      archivePrefix={arXiv},
      primaryClass={cs.LG},
      url={https://arxiv.org/abs/2504.16506}, 
}


%




\end{document}